\pdfoutput=1

\documentclass[11pt]{article}
\usepackage[]{ACL2023}

\usepackage{times}
\usepackage{latexsym}
\usepackage{amsmath}
\usepackage{multirow}
\usepackage{graphicx}
\usepackage{amsfonts}

\usepackage[T1]{fontenc}

\usepackage[utf8]{inputenc}

\usepackage{microtype}

\usepackage{makecell}
\usepackage{graphicx}
\usepackage{appendix}
\usepackage{multirow}
\usepackage{amsfonts,amssymb}
\title{USTC-NELSLIP at SemEval-2023 Task 2: Statistical Construction and Dual Adaptation of Gazetteer for Multilingual Complex NER}

\author{Jun-Yu Ma\textsuperscript{1}, Jia-Chen Gu\textsuperscript{1}, Jiajun Qi\textsuperscript{1},
Zhen-Hua Ling\textsuperscript{1}, Quan Liu\textsuperscript{2} \and Xiaoyi Zhao\textsuperscript{3} \\
\textsuperscript{1}National Engineering Research Center of Speech and Language Information Processing,\\
University of Science and Technology of China\\ 
\textsuperscript{2}State Key Laboratory of Cognitive Intelligence, iFLYTEK Research \\
\textsuperscript{3}Communication University of China\\
{\tt \{mjy1999,jiajun97\}@mail.ustc.edu.cn},
{\tt \{gujc,zhling\}@ustc.edu.cn}, \\
{\tt quanliu@iflytek.com},
{\tt zxy2602021@163.com}
}

\newcommand{\MODELNAME}{SCDAG}
\newcommand{\FULLMODELNAME}{Statistical Construction and Dual Adaptation of Gazetteer}

\begin{document}
\maketitle
\begin{abstract}


This paper describes the system developed by the USTC-NELSLIP team for SemEval-2023 Task 2 Multilingual Complex Named Entity Recognition (MultiCoNER II).
A method named Statistical Construction and Dual Adaptation of Gazetteer (SCDAG) is proposed for Multilingual Complex NER.
The method first utilizes a statistics-based approach to construct a gazetteer.
Secondly, 
the representations of gazetteer networks and language models are adapted by minimizing the KL divergence between them at both the sentence-level and entity-level.
Finally, these two networks are then integrated for supervised named entity recognition (NER) training.
The proposed method is applied to XLM-R with a gazetteer built from Wikidata, and shows great generalization ability across different tracks. 
Experimental results and detailed analysis verify the effectiveness of the proposed method.
The official results show that our system ranked \textbf{1st} on one track (Hindi) in this task.
\end{abstract}

\section{Introduction}
\label{intro}
Named Entity Recognition (NER) is a fundamental and important natural language processing (NLP) task, which aims at finding entities and recognizing their type in a text sequence.
Recently, deep neural networks have achieved great performance on simple NER with abundant labeled data~\cite{DBLP:conf/acl/YeL18,DBLP:conf/emnlp/JiaSYZ20,DBLP:conf/semeval/ChenMQGLL22}.
In practical and open-domain settings, it is difficult for machines to process complex and fine-grained named entities~\cite{DBLP:journals/corr/AshwiniC14,multiconer2-report}.
For example, ``\emph{The Old Man and the Sea}'' is the title of a movie as well as a book, which has  different categories in different contexts and cannot be recognized easily by present NER systems.
This issue may become even more serious in multilingual settings~\cite{DBLP:conf/sigir/FetahuFRM21}.
However, it has not received sufficient attention from the research community.
To alleviate the issue, SemEval-2023 Task 2 \cite{multiconer2-report} formulates this task which 
challenges participants to develop NER systems for 12 languages (English, Spanish, Swedish, Ukrainian, Portuguese, French, Farsi, German, Chinese, Hindi, Bangla and Italian),
focusing on recognizing semantically complex and fine-grained entities in short and low-context settings.
Each language constitutes a single track, while Multilingual is added as Track 13.
The datasets~\cite{multiconer2-data} mainly contain sentences from three domains: Wikipedia, web questions and  search queries.
Besides, simulated errors are added to the test set to make the task more realistic and difficult.


Recent studies have found that integrating external knowledge or gazetteers into neural architectures is effective in solving this problem \cite{DBLP:conf/acl/LiuYL19,DBLP:conf/acl/RijhwaniZNC20,DBLP:conf/naacl/MengFRM21}. 
For example, the two representations respectively from a language model and a gazetteer network are integrated as one representation, which is further fed into a classifier such as a conditional random field (CRF) \cite{DBLP:conf/icml/LaffertyMP01}.
However, for the fine-grained entities, due to the closer semantic distance between these entity categories, the coverage rate of the constructed entity gazetteer is difficult to improve.
Besides, the interaction between the two representations in previous work~\cite{DBLP:conf/semeval/ChenMQGLL22} only focus on sentence-level, which ignore the entity-level representation gap between them and lead to information loss.


In this paper, we propose a method named {\FULLMODELNAME{}} (\MODELNAME{}) for Multilingual Complex NER based on GAIN~\cite{DBLP:conf/semeval/ChenMQGLL22}.
 Firstly, based on Wikipedia of the 12 languages, we build a multilingual gazetteer to search for the entities in input sentence.
 Different from GAIN, we use a statistics-based approach to maximize the coverage of the gazetteer.
Afterwards, the {\MODELNAME{}} adopts a two-stage training strategy to dually adapt the gazetteer network to the language model.
During the first training stage, the parameters of a language model are fixed.
Then a sentence and its annotation are fed into the two networks separately.
The representations of
gazetteer networks and language models are
adapted by minimizing the KL divergence between them at the sentence-level and entity-level.
This process helps the gazetteer network understand the meaning of NER tags
and strengthen the model adaptation ability to NER.
A gazetteer is applied to sentences to generate pseudo tags which are fed into the two pre-trained networks separately in the second stage.
Finally, the two output representations are integrated for classifying.

The proposed method achieves great improvements on the validation set \cite{multiconer2-data} of SemEval-2023 Task 2 compared to baseline models with gazetteers.
Ensemble models are used for all thirteen tracks in the final test phase, and our system officially ranked \textbf{1st} on one track (Hindi).
The outstanding performance demonstrates the effectiveness of our method.
To facilitate the reproduction of our results, the code is available at \url{https://github.com/mjy1111/SCDAG}. 
\begin{figure*}[ht]
\includegraphics[width=\textwidth]{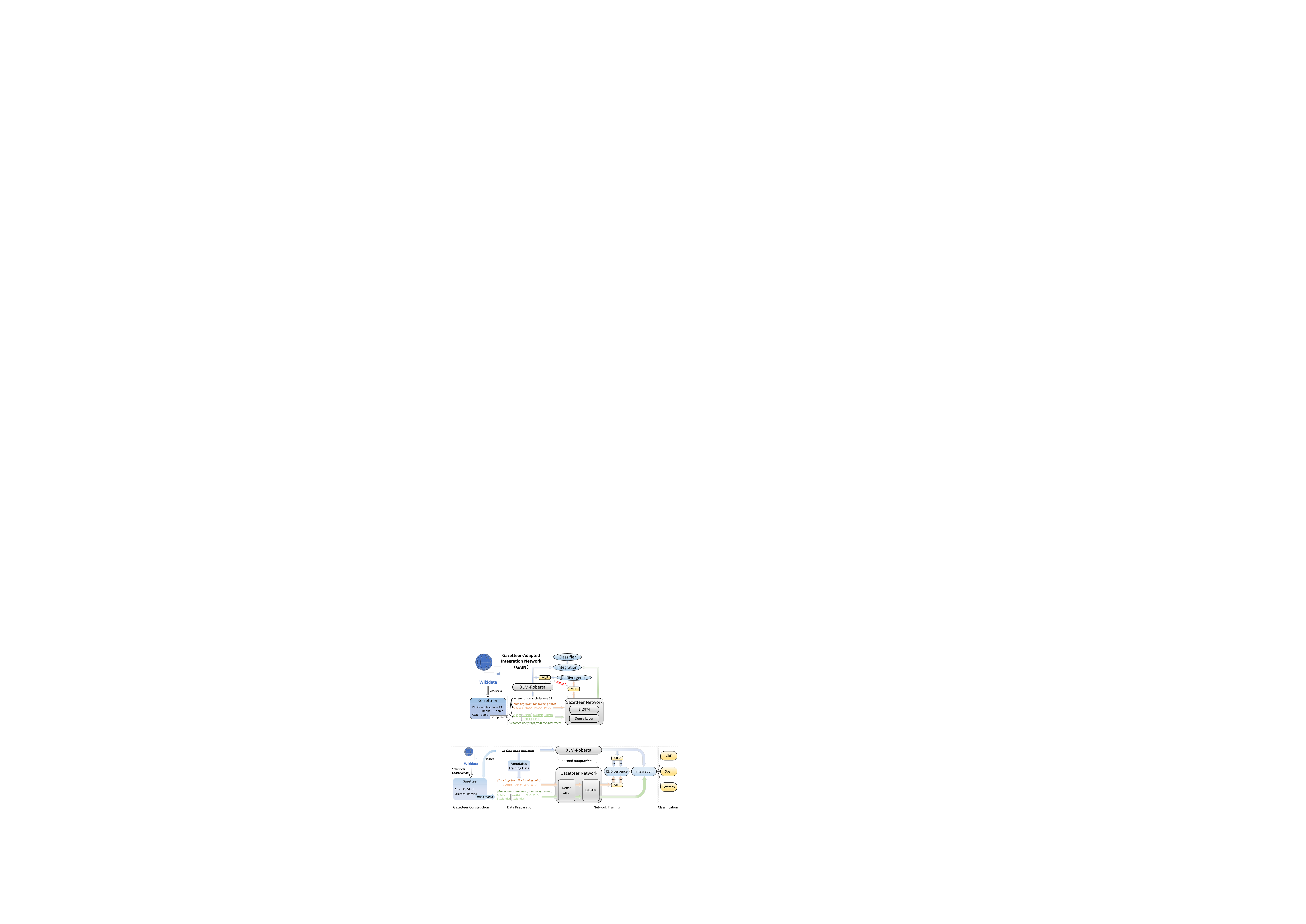}
\caption{The overall structure of the proposed system.
``S'' and ``E'' are the logits distributions of the whole sentence and entities, reflecting the sentence-level and entity-level adaptation, respectively.}  \label{fig1}
\end{figure*}




\section{Related Work}
NER has a lot of applications in various domains and languages.
Recently, with the introduction of contextual pre-trained models, such as BERT, ROBERTA~\cite{DBLP:conf/emnlp/DelobelleWB20} and XLM-R~\cite{DBLP:conf/acl/ConneauKGCWGGOZ20}, the performance of NER systems has been significantly improved.
These models are trained on large-scale unlabeled data such as Wikipedia, which can significantly improve the
contextual representations abilities.

 SemEval-2023 Task 2 is a continuation
of the multilingual NER task started in 2022~\cite{multiconer-report}.
There are many challenges that can make NER extremely difficult.
 In \citet{DBLP:conf/naacl/MengFRM21}, they explain
that named entity recognition is especially difficult
in situations with low-context or in scenarios where
the named entities are exceptionally complex and ambiguous.
Another work has extended this to multilingual and code-mixed settings~\cite{DBLP:conf/sigir/FetahuFRM21}.
These are the key challenges of the 2022 datasets~\cite{multiconer-data} and we have participated in this competition~\cite{DBLP:conf/semeval/ChenMQGLL22}.
Besides, NER requires abundant well-annotated data, which is too expensive in low-resource languages~\cite{DBLP:conf/emnlp/MaCGLGL0L22}.

Lots of methods are proposed to improve NER performance and a general discovery is that the use of external knowledge bases is very effective.
\citet{DBLP:conf/acl/WangJBWHHT20} retrieve related contexts
from a search engine as external contexts of the
inputs to take advantage of long-range
dependencies for entity disambiguation and successfully achieve state-of-the-art performance across
multiple datasets.
\citet{DBLP:conf/naacl/MengFRM21} recognize the importance
of gazetteer resources, even in the case of state-of-the-art systems making use of pre-trained models. 
They propose a Contextual Gazetteer Representation encoder, combined with a novel
Mixture-of-Expert (MoE) gating network to conditionally utilize gazetteer information.
\citet{DBLP:conf/sigir/FetahuFRM21} employ multilingual
gazetteers embedded with transformer models in an MoE
approach to improve the recognition of entities in
code-mixed web queries, where entities are in a different language from the rest of the query.




\section{Data} \label{data}
The MultiCoNER dataset~\cite{multiconer2-data} is provided in
a column-based format and divided into the training,
development and testing sets.
The text is lowercase with the named entity annotation in BIO format~\cite{DBLP:conf/conll/SangM03}. The first token (or the single
token) of an entity contains the ``B-'' prefix. Other
entity tokens (in the case of multi-token entities)
start with an ``I-'' prefix, while non-entity tokens
are denoted with ``O''.
It consists of 6 coarse-grained entity types and 33 fine-grained entity types, and the coarse to fine level mapping of the tags is as follows:
Location (Facility, OtherLOC, HumanSettlement, Station), Creative Work (VisualWork, MusicalWork, WrittenWork, ArtWork, Software), Group (MusicalGRP, PublicCORP, PrivateCORP, AerospaceManufacturer, SportsGRP, CarManufacturer, ORG), 
Person (Scientist, Artist, Athlete, Politician, Cleric, SportsManager, OtherPER),
Product (Clothing, Vehicle, Food, Drink, OtherPROD),
Medical (Medication/Vaccine, MedicalProcedure, AnatomicalStructure, Symptom, Disease).

Two data augmentation methods are used following~\cite{DBLP:conf/semeval/ChenMQGLL22}.
For the basic training set provided officially, an entity replacement strategy is adopted using our own gazetteer to construct a data-augmented set. 
This part of data is called ``data-wiki'', which mainly consists of rich-context sentences.
In order to improve the performance of our models on low-context instances, a set of annotated sentences are generated from the MS-MARCO QnA corpus (V2.1) \cite{DBLP:conf/nips/NguyenRSGTMD16} and the ORCAS dataset \cite{DBLP:conf/cikm/CraswellCMYB20}, which are mentioned in \citet{DBLP:conf/naacl/MengFRM21}.
Our trained models and existing NER systems (e.g., spacy) are applied to identify entities in these corpora, and only templates identically recognized by all models are reserved.
Then 3,379 English templates for MS-MARCO and 11,754 English templates for ORCAS are obtained.
Next, we slot the templates by our own gazetteer and translate them to the other 11 languages.
This part of data is called ``data-query''.
Finally, the constructed data is used together with the official data for training.



\section{Methodology}

This study focuses on making better use of external entity knowledge.
To describe our system clearly, in this section, we first introduce three basic mainstream NER systems used.
Then we show the process of constructing a gazetteer from Wikipedia using a statistics-based approach to maximize the coverage rate, and how the gazetteer representation is generated and utilized.
Finally, we illustrate the dual adaptation between gazetteer network and language model.
The overall structure of the proposed system is shown in Figure~\ref{fig1}.

\subsection{Basic NER Systems}\label{sect: backend}
We mainly use the XLM-RoBERTa large \cite{DBLP:conf/acl/ConneauKGCWGGOZ20} as the pre-trained language model, which is a widely used encoder.
Generated by feeding a sentence into the encoder, the representation is then input to different classifiers.
Three mainstream NER backend classifiers are adopted: Softmax \cite{DBLP:conf/naacl/DevlinCLT19} and CRF \cite{DBLP:journals/corr/HuangXY15} are classic sequential labeling methods that predict the tag of each token, and Span \cite{DBLP:conf/acl/YuBP20} is a segment-based method that predicts the start and the end of an entity separately.




\begin{table}[]
\begin{center}
\begin{tabular}{l|c|c|c}
\Xhline{1.5pt}                                
 Lang.          & Total Num.                  & Average &  Average-stat \\\Xhline{1pt}

EN &3035,896 & 32.52\% & 36.76\% \\
ZH &940,614 & 27.83\% & 32.59\% \\
HI &80,588 & 43.86\% & 46.55\% \\
BN &115,759 & 43.20\% & 45.91\% \\
ES &1334,659 & 40.94\% & 42.34\% \\
DE &1686,065 & 30.44\% & 32.27\% \\
FA &460,283 & 40.81\% & 43.61\% \\
FR &785,379 & 27.66\% & 30.71\% \\
IT &243,768 & 33.49\% & 35.86\% \\
PT &124,532 & 19.78\% & 20.99\% \\
SV &79,137 & 17.67\% & 21.53\% \\
UK &101,986 & 30.34\% & 35.38\% \\
MULTI &6269,437 & 31.29\% & 35.73\% \\

\Xhline{1.5pt}
\end{tabular}
\end{center}
\caption{The metrics of our gazetteer in detail. The Total Num. column means the accurate number of entries in the gazetteer for each track. Numbers with \% denote the coverage rates to entities in the training and validation set. The Average means coverage rate of the gazetteer with a manual one-to-one matching following~\citet{DBLP:conf/semeval/ChenMQGLL22} and the Average-stat is the coverage rate after adding the statistics-based approach.}\label{tabx}
\end{table}

\subsection{The Gazetteer}\label{sect: gazetteer}
It's difficult to process complex and fine-grained entities only relying on the language model itself \cite{DBLP:journals/corr/AshwiniC14}. 
To integrate external entity knowledge, we first need to build a large gazetteer matching the taxonomy, then we have to consider how to fuse the gazetteer information with the semantic information from the language model.

\subsubsection{Statistical Construction}
Our gazetteer is built based on Wikidata.
Wikidata is a free and open knowledge base.
Every entity of Wikidata has a page consisting of a label, several aliases, descriptions, and one or more entity types.
We use entity type annotated by Wikidata to construct the gazetteer.
For example, ``\emph{Da Vinci}'' can be annotated as a researcher or a well-known artist in Wikidata.
Thus, according to the entity definition of this competition, the word ``\emph{Da Vinci}'' is given both scientist and artist labels.

In the previous work~\cite{DBLP:conf/semeval/ChenMQGLL22}, to construct a gazetteer fit to the data of this task, firstly every entity of the training set is searched in Wikidata. Then all the entity types returned are mapped to the NER taxonomy with 6 labels manually.
Next, all Wikidata entities stored in these entity types can be added to the 6 labels gazetteer separately.
However, since labels are fine-grained and some labels are semantically similar in this work, the cost of manual matching is too high and just adding the returned entity types to a certain label gazetteer will result in low coverage of the gazetteer.
Therefore, we counted the coverage of the Wikidata entities contained in each returned entity type on each label of the training set, and added this entity type to the top two labels gazetteer with the highest coverage.

In this way, a multilingual gazetteer is constructed that contains entities from 70K to 3M for each language.
The gazetteer approximately has a coverage rate of 35 percent on entities in the training and validation set.
To validate the effectiveness of this method, we also use a manual one-to-one matching same as~\citet{DBLP:conf/semeval/ChenMQGLL22}.
Basic information about our gazetteer is shown in Table ~\ref{tabx}.
The coverage rate is calculated as the number of entities both appearing in the official data and our gazetteer divided by the total number of entities in the official data.
It shows that with the statistic-based approach, the coverage rate of the gazetteer gains a significant improvement.

\begin{table}[t]
    \centering
    \setlength{\tabcolsep}{1.5mm}
    \begin{tabular}{l|ccccc}
    \Xhline{1.5pt}
         \footnotesize{Words} &\scriptsize{O} &\scriptsize{B-Food} &\scriptsize{I-Food} &\scriptsize{B-OtherPROD} &\scriptsize{I-OtherPROD}  \\\hline
         where &1 &0 &0 &0 &0 \\\hline
         to &1 &0 &0 &0 &0 \\\hline
         buy &1 &0 &0 &0 &0 \\\hline
         apple &0 &1 &0 &1 &0 \\\hline
         iphone &0 &0 &0 &1 &1 \\\hline
         14 &0 &0 &0 &0 &1 \\\Xhline{1.5pt}
    \end{tabular}
    \caption{Example of the one-hot representation for a searched sentence. The rest labels are all zero.}
    \label{tab2}
\end{table}

\subsubsection{Application}\label{usegaze}
A search tree is constructed for string matching firstly to apply the gazetteer to a sentence.
Once a sentence is fed into the search tree, a maximum length matching algorithm will be conducted, and a 67-dimension one-hot vector for each token will be generated.
Take the sentence ``\emph{where to buy apple iphone 14}'' for example.
By string matching with the gazetteer, ``\emph{apple iphone 14}'', ``\emph{iphone 14}'' and ``\emph{apple}'' are found in the OtherPROD gazetteer, while ``\emph{apple}'' is also found in the Food gazetteer.
Then a 67-dimension one-hot vector will be generated for every word as shown in Table~\ref{tab2}.

Denote one sentence as \(\mathbf{w}=(w_1,w_2,...w_M)\) where \(M\) is the length of the sentence and \(w_i\) is the \(i^{th}\) word.
By feeding \(\mathbf{w}\) into the encoder such as the XLM-RoBERTa large, a semantic representation \(\mathbf{e}\in \mathbb{R}^{N\times D}\) is obtained, where N is the length of subword tokens and D is the hidden size.
At the same time, the one-hot vector generated from the search tree is fed into a gazetteer network consisting of a dense layer and a BiLSTM.
To match the hidden size of the language model, the output embedding \(\mathbf{g}\) has the same size as \(\mathbf{e}\).
Noting that the value of each word in gazetteer is assigned to the first subword, and the other subwords are 0.

Then we use two ways to integrate \(\mathbf{e}\) and \(\mathbf{g}\).
One way is to concatenate them on each token, another way is to get the weighted summation of them by setting a trainable parameter \(\lambda\in \mathbb{R}^{N\times D}\). The final representation is fed into the backend classifier for supervised NER training.


\subsection{Dual Adaptation}\label{GAIN}
\citet{DBLP:conf/semeval/ChenMQGLL22} have found that only conducting the normal training process above is not enough. 
Since the encoder XLM-R large and the gazetteer network BiLSTM are almost isolating each other during the whole training, almost no semantic information can be gained explicitly by the classic gazetteer network.

In this paper, the  
dual adaptation method is proposed with a two-stage training strategy to interact the representations of gazetteer network and language model at both sentence-level and entity-level.
In the first stage, 
an one-hot vector is constructed just based on the true tags in training set for each sentence.
A gazetteer representation \(\mathbf{g}_{r}\in \mathbb{R}^{N\times D}\) is obtained after passing the vector through the gazetteer network.
Then the parameters of the language model are fixed, and the sentence \(\mathbf{w} \) is fed into it to get a semantic representation \(\mathbf{s}\).
\(\{\mathbf{g}_{r}, \mathbf{s}\}\) are projected to \(\{\mathbf{g}^{t}_{r},\mathbf{s}^{t}\} \in \mathbb{R}^{N\times 67}\) by two separate linear layers, where the semantic meaning is transferred to the tags meaning as a kind of logits distribution.
Meantime, all entity logits are connected and denoted as \(\{\mathbf{g}^{e}_{r},\mathbf{s}^{e}\} \in \mathbb{R}^{E\times 67}\).
Then the sentence-level and entity-level adaptation are respectively implemented by the designed losses:


\begin{align}\label{eq1}
    \emph{L}_{1}&={\rm KL}(sg(\mathbf{g}^{t}_{r})||\mathbf{s}^{t})+{\rm KL}(sg(\mathbf{s}^{t})||\mathbf{g}^{t}_{r}), \\
    \emph{L}_{2}&={\rm KL}(sg(\mathbf{g}^{e}_{r})||\mathbf{s}^{e})+{\rm KL}(sg(\mathbf{s}^{e})||\mathbf{g}^{e}_{r}),
    \label{eq2}
\end{align}


\noindent where \({\rm KL}(\cdot)\) is the KL divergence calculation and \(sg(\cdot)\) operation is used to stop back-propagating gradients, which
is also employed in \citet{DBLP:conf/acl/JiangHCLGZ20,DBLP:journals/corr/abs-2004-08994}.
The loss \(\emph{L}_{1}\) encourages the distributions \(\mathbf{g}^{t}_{r}\) and \(\mathbf{s}^{t}\) to
approximate each other to enhance the two networks as a whole at the sentence-level.
The loss \(\emph{L}_{2}\)
utilizes entities
 to strengthen
the model adaptation ability to NER.
In the second stage, all the parameters are trained with a gazetteer.
As illustrated in Section \ref{usegaze}, a gazetteer representation \(\mathbf{g}\) is generated from the search tree and the gazetteer network BiLSTM. Next, an ordinary fusion method is applied to \(\mathbf{g}\) and \(\mathbf{s}\) to get an integration representation, which is then fed into the backend classifier to compute a conventional loss with true tags \({\rm T}\). This supervised training goal is implemented by the loss \(\emph{L}_{3}\):

\begin{equation}
\emph{L}_{3} = {\rm Classifier}(f(\mathbf{g},\mathbf{s}),{\rm T}),
\end{equation}
\noindent where \(f(\cdot)\) denotes ordinary integration methods like concatenation or weighted summation. \({\rm Classifier}(\cdot)\) represents one of the three mainstream backend classifiers mentioned in Section \ref{sect: backend}. 
It is worth noting that in Eq.~(\ref{eq2}), we use the true entities in training set for the whole training.
During the second-stage training, a multitask learning goal is conducted shown as:

\begin{equation}
\emph{L}_{4} = \alpha(\emph{L}_{1} + \emph{L}_{2})+\emph{L}_{3},
\end{equation}

\noindent where \(\alpha\) is a hyperparameter control the importance of gazetteer and is manually set for different fusion or backend methods.

\section{Experiments}
\label{sec:results}

\begin{table*}[]
\footnotesize
    \centering
    \setlength{\tabcolsep}{0.4mm}{
    \begin{tabular}{c|l|cccccccccccc}
    \Xhline{1.5pt}
\multicolumn{1}{l|}{Domain} & Metrics\textbackslash{}Lang & EN             & ZH             & HI             & BN             & ES             & DE             & FA             & FR             & IT             & PT             & SV &UK                  \\\hline
\multirow{9}{*}{overall}  & f-macro@F1 &0.7215	&0.6657	&0.8214	&0.8059	&0.7444	&0.7871	&0.6885	&0.7425	&0.7570	&0.7126	&0.7547 &0.7437
\\
                            & f-macro@P &0.7464	&0.6695	&0.8306	&0.8140	&0.7581	&0.7773	&0.6827	&0.7573	&0.7608	&0.7076	&0.7444 &0.7384
\\
                            & f-macro@R &0.6996	&0.6686	&0.8319	&0.8132	&0.7355	&0.8026	&0.7093	&0.7302	&0.7647	&0.7402	&0.7811 &0.7688
\\
                            & c-macro@F1 &0.8203	&0.7948	&0.9025	&0.9002	&0.7355	&0.8898	&0.7796	&0.8306	&0.8568	&0.8468	&0.8666 &0.8550
\\
                            & c-macro@P &0.8434	&0.8103	&0.9119	&0.9119	&0.8538	&0.8919	&0.7866	&0.8420	&0.8560	&0.8630	&0.8702 &0.8605
\\
                            & c-macro@R &0.7989	&0.7800	&0.8938	&0.8839	&0.7198	&0.8879	&0.7730	&0.8197	&0.8488	&0.8401	&0.8630 &0.8595
\\
                           & TRUE                    &377,805	&27,999	&23,199	&25,013	&356,374	&28,877	&312,115	&398,195	&397,222	&340,752	&361,159 &315,374

 \\
                           & PRED                    &361,714	&26,920	&22,790	&24,367	&344,583	&28,658	&309,831	&389,442	&391,749	&335,472	&356,705 &311,695

 \\
                           & RECALLED                &328,276	&23,840	&21,468	&23,116	&313,180	&26,924	&255,270	&351,401	&362,917	&306,415	&331,177 &281,135

 \\
\hline
\multirow{7}{*}{LOC}   & macro@F1                    &0.8806	&0.8422	&0.9410	&0.9161	&0.8638	&0.9227	&0.8009	&0.8542	&0.8784	&0.8856	&0.9292	&0.8941
 \\
                           & macro@P                     &0.8931	&0.8636	&0.9440	&0.9531	&0.8811	&0.9237	&0.7960	&0.8659	&0.8876	&0.8970	&0.9428	&0.9007
 \\
                           & macro@R                    &0.8685	&0.8219	&0.9380	&0.9470	&0.8473	&0.9218	&0.8059	&0.8429	&0.8695	&0.8744	&0.9160	&0.8875
 \\
                           & F1@Facility                      &0.7462	&0.7002	&0.8108	&0.8515	&0.7193	&0.7969	&0.7050	&0.7457	&0.7959	&0.7417	&0.8069	&0.7500
 \\
                           & F1@otherloc                     &0.7585	&0.5781	&0.8445	&0.8444	&0.6352	&0.7134	&0.5514	&0.6773	&0.6571	&0.7883	&0.9032	&0.7380
 \\
                           & F1@HS                     &0.9125	&0.8509	&0.9473	&0.9566	&0.8902	&0.9386	&0.8134	&0.8687	&0.8963	&0.9053	&0.9483	&0.9145
 \\
                           & F1@Station                     &0.8579	&0.8482	&0.9298	&0.9100	&0.8036	&0.8464	&0.8629	&0.8137	&0.8055	&0.8121	&0.8413	&0.8037
 \\
\hline
\multirow{7}{*}{CW}    & macro@F1                    & 0.8175	&0.7763	&0.8697	&0.9161	&0.8383	&0.8819	&0.7973	&0.8393	&0.9014	&0.8463	&0.8285	&0.8187
 \\
                           & macro@P                  &0.8519	&0.7965	&0.9139	&0.9046	&0.8629	&0.8819	&0.7973	&0.8570	&0.9130	&0.8651	&0.8412	&0.8210
 \\
                           & macro@R                   &0.7858	&0.7572	&0.8295	&0.8176	&0.8151	&0.8687	&0.7789	&0.8224	&0.8902	&0.8283	&0.8162	&0.8163
 \\
                           & F1@Visual                      &0.8190	&0.7453	&0.8947	&0.8674	&0.8192	&0.8355	&0.8430	&0.8833	&0.9312	&0.8170	&0.8475	&0.8235
 \\
                           & F1@Musical                     &0.8181	&0.6215	&0.6281	&0.8079	&0.7784	&0.8490	&0.7161	&0.7668	&0.8676	&0.8180	&0.8175	&0.7381
 \\
                           & F1@Written                    &0.7844	&0.7425	&0.8919	&0.8834	&0.7820	&0.8542	&0.6820	&0.7692	&0.7539	&0.7447	&0.7414	&0.7693
 \\
                           & F1@Art                      &0.6859	&0.5305	&0.4692	&0.3678	&0.5917	&0.8079	&0.2640	&0.7025	&0.7363	&0.2613	&0.4631	&0.5047
 \\
                           & F1@Software                     &0.8264	&0.7183	&0.9332	&0.9357	&0.8694	&0.8841	&0.7722	&0.8131	&0.8393	&0.8502	&0.8657	&0.8748
 \\
\hline
\multirow{10}{*}{GRP}      & macro@F1                  &0.8202	&0.8076	&0.9443	&0.9353	&0.8494	&0.9005	&0.7967	&0.8311	&0.8617	&0.8589	&0.8621	&0.8751
\\
                           & macro@P                   &0.8581	&0.8304	&0.9558	&0.9500	&0.8846	&0.9042	&0.8109	&0.8403	&0.8712	&0.8680	&0.8600	&0.8821
 \\
                           & macro@R                    &0.7854	&0.7860	&0.9330	&0.9211	&0.8170	&0.8968	&0.7829	&0.8222	&0.8524	&0.8499	&0.8641	&0.8682
 \\
                           & F1@Musical                   &0.8259	&0.7349	&0.9516	&0.9443	&0.8396	&0.8627	&0.8253	&0.8235	&0.8688	&0.8394	&0.8621	&0.8873
  \\
                           & F1@Public                   &0.7648	&0.6268	&0.9140	&0.8849	&0.8267	&0.7883	&0.7706	&0.7808	&0.8291	&0.8540	&0.7972	&0.8691
 \\
                           & F1@Private                &0.7089	&0.7395	&0.8834	&0.9644	&0.6824	&0.8848	&0.6358	&0.7961	&0.5692	&0.0508	&0.7066	&0.4537
 \\
                           & F1@AM                     &0.7692	&0.7040	&0.4715	&0.4615	&0.7349	&0.9037	&0.8832	&0.7639	&0.6431	&0.4832	&0.4832	&0.5771
 \\
                           & F1@Sports                      &0.8872	&0.8647	&0.9729	&0.9651	&0.871	&0.9363	&0.8843	&0.8554	&0.8777	&0.8801	&0.9089	&0.9123
 \\
                           & F1@CM                    &0.7988	&0.7083	&0.9246	&0.9195	&0.8307	&0.8449	&0.7859	&0.7907	&0.8055	&0.8175	&0.8003	&0.8561
 \\
                           & F1@ORG                    &0.7575	&0.7239	&0.9256	&0.9157	&0.7594	&0.8156	&0.6965	&0.7165	&0.7280	&0.7617	&0.7691	&0.7898
 \\ \hline
\multirow{10}{*}{PER}    & macro@F1                   &0.9419	&0.8942	&0.9287	&0.9294	&0.9463	&0.9425	&0.8566	&0.9483	&0.9625	&0.9472	&0.9586	&0.9481
 \\
                           & macro@P                    &0.9450	&0.9062	&0.9313	&0.9400	&0.9486	&0.9451	&0.8515	&0.9498	&0.9626	&0.9465	&0.9584	&0.9505
 \\
                           & macro@R                    &0.9388	&0.8825	&0.9261	&0.9191	&0.9440	&0.9399	&0.8617	&0.9467	&0.9624	&0.9479	&0.9588	&0.9457
 \\
                           & F1@Scientist                     &0.5759	&0.4085	&0.6448	&0.5104	&0.5871	&0.4660	&0.4612	&0.5656	&0.5786	&0.5281	&0.5315	&0.5686
 \\
                           & F1@Artist                      &0.8016	&0.7471	&0.7834	&0.7762	&0.8307	&0.7992	&0.7887	&0.8451	&0.8929	&0.8443	&0.8241	&0.8033
 \\
                           & F1@Athlete                     &0.8150	&0.7691	&0.8428	&0.7813	&0.8122	&0.8135	&0.7011	&0.8275	&0.8938	&0.8035	&0.8215	&0.8467
 \\
                           & F1@Politician                     &0.7074	&0.5575	&0.7581	&0.7086	&0.6888	&0.6533	&0.6661	&0.6986	&0.7087	&0.7236	&0.7345	&0.6529
 \\
                           & F1@Cleric                      &0.7175	&0.4516	&0.8395	&0.7388	&0.6859	&0.5956	&0.6320	&0.6946	&0.7756	&0.7421	&0.6877	&0.6705
 \\
                           & F1@SM                    &0.6581	&0.5174	&0.6419	&0.6440	&0.6788	&0.5959	&0.6432	&0.6456	&0.7235	&0.6517	&0.5983	&0.6716
 \\
                           & F1@otherper                      &0.5418	&0.4986	&0.5959	&0.5734	&0.6127	&0.5818	&0.5137	&0.5875	&0.6095	&0.6145	&0.5985	&0.6093
 \\
\hline
\multirow{8}{*}{PROD}    & macro@F1                    &0.6908	&0.6972	&0.8302	&0.8112	&0.7355	&0.8134	&0.7073	&0.7302	&0.7448	&0.7655	&0.7939	&0.7743
 \\
                           & macro@P                    &0.7262	&0.6977	&0.8304	&0.8139	&0.7520	&0.8071	&0.7096	&0.7495	&0.7576	&0.7676	&0.7931	&0.7870
 \\
                           & macro@R                    &0.6587	&0.6966	&0.8299	&0.8084	&0.7198	&0.8198	&0.7050	&0.7118	&0.7323	&0.7635	&0.7948	&0.7619
 \\
                           & F1@Clothing                     &0.6915 &0.6267	&0.8249	&0.5778	&0.6713	&0.7703	&0.4948	&0.6789	&0.6574	&0.5989	&0.7198	&0.6740
 \\
                           & F1@Vehicle                     &0.5418	&0.6739	&0.9029	&0.8543	&0.7099	&0.7704	&0.6483	&0.6455	&0.6751	&0.6911	&0.7187	&0.7301
 \\
                           & F1@Food                    &0.6671	&0.6973	&0.8181	&0.7843	&0.6571	&0.7334	&0.6544	&0.6338	&0.6484	&0.6804	&0.7164	&0.7161
 \\
                           & F1@Drink                      &0.6820	&0.5319	&0.8651	&0.8651	&0.7321	&0.7500	&0.6649	&0.6722	&0.7234	&0.7245	&0.7675	&0.7538
 \\
                           & F1@otherprod                   &0.6602	&0.6340	&0.7823	&0.7733	&0.6573	&0.7568	&0.6844	&0.6826	&0.6871	&0.7577	&0.7563	&0.7229
 \\
\hline
\multirow{8}{*}{MED}    & macro@F1                    &0.7707	&0.7513	&0.9010	&0.9045	&0.7899	&0.8779	&0.7190	&0.7808	&0.7919	&0.7775	&0.8271	&0.8197
 \\
                           & macro@P                    &0.7858	&0.7671	&0.8961	&0.8875	&0.7938	&0.8758	&0.7352	&0.7895	&0.7978	&0.7784	&0.8259	&0.8219
 \\
                           & macro@R                   &0.7561	&0.7361	&0.9060	&0.9221	&0.7860	&0.8801	&0.7036	&0.7722	&0.7861	&0.7766	&0.8282	&0.8175
 \\
                           & F1@Medv                    &0.8078	&0.7403	&0.9122	&0.9045	&0.7987	&0.8787	&0.7837	&0.7980	&0.8121	&0.8105	&0.8183	&0.8571
 \\
                           & F1@Medp                  &0.7234	&0.6955	&0.8754	&0.8885	&0.7440	&0.8362	&0.6923	&0.7405	&0.7525	&0.7427	&0.7494	&0.7182
 \\
                           & F1@Ans                   &0.7579	&0.7276	&0.8785	&0.9178	&0.7729	&0.8536	&0.6458	&0.7217	&0.7567	&0.7321	&0.8246	&0.8203
 \\
                           & F1@Symptom                     &0.6783	&0.5200	&0.8464	&0.9009	&0.6859	&0.6818	&0.6426	&0.7269	&0.7077	&0.6206	&0.6535	&0.6674
 \\
                           & F1@Disease                     &0.7761	&0.7345	&0.9006	&0.9157	&0.7871	&0.8747	&0.7132	&0.7695	&0.7727	&0.7936	&0.8235	&0.7961
 \\
\hline

\Xhline{1.5pt}
\end{tabular}}
    \caption{All detailed results of the official test set on monolingual tracks.``f-'' and ``c'' referred the fine-grained and coarse-grained respectively.
    Due to limited spaces, some class labels were abbreviated. ``Work'', ``GRP'' and ``CORP'' were omitted in all labels.
    ``HS'' was HumanSettlement, ``AM'' was AerospaceManufacturer, ``CM'' was CarManufacturer, ``SM'' was SportsManager, ``Medv'' was Medication/Vaccine, ``Medp'' was MedicalProcedure and ``Ans'' was AnatomicalStructure.
    }
    \label{taball}
\end{table*}

\begin{table*}[]
\footnotesize
    \centering
    \setlength{\tabcolsep}{1.6mm}{
    \begin{tabular}{clcccccccccccc}
    \Xhline{1pt}
\multicolumn{1}{l}{Method}                               & Classifier & EN             & ZH             & HI             & BN             & ES             & DE             & FA             & FR             & IT             & PT             & SV             & UK            \\\hline
\multirow{3}{*}{Base}                                  & CRF       & 0.682          & 0.733          & 0.836          & 0.849          & 0.710           & 0.737          & 0.681          & 0.692          & 0.732          & 0.712          & 0.713          & 0.704          \\

 & Softmax       & 0.671          & 0.725          & 0.828          & 0.834          & 0.702           & 0.733          & 0.672          & 0.687          & 0.729          & 0.701          & 0.716          & 0.709          \\

  & Span       & 0.691          & 0.749          & 0.840          & 0.866          & 0.725           & 0.758          & 0.683          & 0.707          & 0.761          & 0.738          & 0.749          & 0.729 
        
       \\\hline
\multirow{3}{*}{Integration}                             & CRF       & 0.708          & 0.753          & 0.845          & 0.876          & 0.743           & 0.771          & 0.687          & 0.712          & 0.773          & 0.747          & 0.766          & 0.742          \\

 & Softmax       & 0.702          & 0.746          & 0.839          & 0.862          & 0.738           & 0.763          & 0.681          & 0.704          & 0.761          & 0.734          & 0.753          & 0.731          \\

  & Span       & 0.719          & \textbf{0.761}          & 0.851          & 0.880          & 0.749           & 0.778          & 0.693          & 0.718          & 0.776          & 0.751          & 0.762          & 0.740
  \\\hline
\multirow{3}{*}{SCDAG}  & CRF       & 0.738          & 0.759          & 0.871          & 0.882           & 0.793          & 0.815          & 0.732          & \textbf{0.740}          & \textbf{0.803}          & 0.771          & 0.773          & \textbf{0.769}          \\
\multicolumn{1}{l}{}                                       & Softmax   & 0.732          & 0.739          & 0.862          & 0.875          & 0.790          & 0.808          & 0.724          & 0.726          & 0.793          & 0.765          & 0.761          & 0.759          \\
\multicolumn{1}{l}{}                                       & Span      & \textbf{0.745}          & 0.754          & \textbf{0.876}          & \textbf{0.885}          & \textbf{0.802}          & \textbf{0.819}          & \textbf{0.737}          & 0.737          & 0.801          & \textbf{0.776}          & \textbf{0.778}          & 0.763          \\\Xhline{1pt}
\end{tabular}}
    \caption{All fine-grained macro-F1 scores on the validation set. Only scores of the concatenation integration method were listed. ``Base'' denoted baseline systems mentioned in Section~\ref{sect: backend}, ``Integration'' was ordinary integration method with the gazetteer mentioned in Section~\ref{sect: gazetteer}, and SCDAG was the proposed method in Section~\ref{GAIN}. }
    \label{tab4}
\end{table*}

\begin{table*}[]
\footnotesize
    \centering
        \setlength{\tabcolsep}{1.5mm}{
\begin{tabular}{l|cccccccccccc}
\Xhline{1pt}
Method & EN             & ZH             & HI             & BN             & ES             & DE             & FA             & FR             & IT             & PT             & SV             & UK    \\\Xhline{0.75pt}
\MODELNAME{}        & \textbf{0.721} & \textbf{0.665} & \textbf{0.821} & \textbf{0.805} & \textbf{0.744}  & \textbf{0.787} & \textbf{0.688} & \textbf{0.742}  & \textbf{0.756} & \textbf{0.712} & \textbf{0.754} & \textbf{0.743} \\\hline
\MODELNAME{} w/o. data-wiki    & 0.682 & 0.617 & 0.804 & 0.787 & 0.708 & 0.773 & 0.665 & 0.711 & 0.723 & 0.691 & 0.714 & 0.726 \\\hline
\MODELNAME{} w/o. data-query      & 0.691 & 0.631 & 0.812 & 0.779 & 0.701 & 0.756 & 0.678 & 0.734 & 0.732 & 0.699 & 0.752 & 0.729 \\\Xhline{1pt}
\end{tabular}}
    \caption{The ablation study on the constructed data in Section \ref{data}.
    Experiments were conducted on the test set.}
    \label{ablation}
\end{table*}








\subsection{Implementation Details}
In this paper, the XLM-R large model was chosen as the encoder for all tracks, which could be found on the HuggingFace Page \cite{DBLP:journals/corr/abs-1910-03771}.
As for hyperparameter, the hidden size was 1024, batch size was 32 and dropout rate was set to 0.1.
The AdamW~\cite{DBLP:conf/iclr/LoshchilovH19} was used as optimizer.
We adopted a learning rate 2e-5 for language models, 2e-4 for gazetteer networks and 2e-5 for classifiers.
The training epoch for first-stage training was 5 and for second-stage training was 20.
The \(\alpha\) for the second stage training was set to 5 for Softmax and Span, 100 for CRF.
All code was implemented in the PyTorch framework\footnote{https://pytorch.org/}.





\subsection{Training Strategy}

In this paper, a 5-fold cross-validation training strategy was also applied in the evaluation and a lot of models had been trained with the {\MODELNAME{}} method using different classifiers.
Firstly, the prepared data ``data-wiki'' and ``data-query'' were split into five pieces, each one was used as the validation set, while the other four pieces were used as the training set.
After obtaining the five best models by this strategy, the logits of them (for Softmax and Span models) were averaged to integrate them as an aggregated model. CRF models had been just voted averagely at the word level.

Finally, the predictions of our best models in different methods were token-voted by setting a weight for each track.
The weight was manually set referring to all scores on the validation set.

\subsection{Official Results}
Our team participated in all 13 tracks and  
the overall fine-grained F1 and per-class performance were reported in Table~\ref{taball}. We also provided the coarse-grained metrics F1.
We ranked \textbf{1st} on the HI track.
As shown in the table, the proposed \MODELNAME{} method significantly improved the performance of recognizing the fine-grained entities.

\subsection{Analysis}

\label{sec:analysis}

\subsubsection{Effectiveness of {\MODELNAME{}}} \label{effect}
To explore the effectiveness of each module in the proposed {\MODELNAME{}} method, a large number of trials were conducted on the official data mentioned in Section~\ref{data}.
All scores under the concatenation integration setting on the validation set were listed in Table~\ref{tab4}.
Compared ``Integration'' with ``Base'', significant improvements were gained by the gazetteer on all tracks.
Compared ``Integration'' with SCDAG, we could find that the dual adaptation and the two-stage training strategy was effective for NER.
Besides, the effect of the CRF method was stronger than that of Softmax, which may be because fine-grained classification was more difficult for linear classifiers.

\subsubsection{Ablation Study on Constructed Data}\label{gradient}
To validate the contribution of the constructed data, the following variants were conducted to perform the ablation study on the test set:
 (1) \MODELNAME{} w/o. data-wiki, which removed the entity replacement strategy.
  (2) \MODELNAME{} w/o. data-query, which removed the templates of MS-MARCO and ORCAS corpus mentioned in Section~\ref{data}.

The results of the ablation experiments were shown in Table~\ref{ablation}.
Some in-depth analysis could be explored:
(1) Compared \MODELNAME{} with \MODELNAME{} w/o. data-wiki, the removal of the entity replacement strategy caused a significant performance drop, especially for EN, ZH, IT, ES, FR, PT and SV.
This was because a noisy subset was held for each language where the sentences were corrupted with noise either on context tokens or entity tokens in the test set~\cite{multiconer2-data}. The ``data-wiki'' enhanced the robustness of the model.
(2) Compared \MODELNAME{} with \MODELNAME{} w/o. data-query, we could see that the removal of the templates caused a significant performance drop, which further demonstrated the importance of introducing short sentence training data for low-context settings. 

\subsubsection{Average-Logits Experiments}

This section explained why we chose to average logits of softmax-based models (for Softmax and Span models) for integrating them as an aggregated model, rather than an average token-vote.
Also, a 5-fold cross-validation training was conducted with the official training data on the basic Softmax method.
Without loss of generality, HI, EN, ZH and FA were chosen to represent different language families.
The results of the official validation set were shown in Table~\ref{taby}.
It was empirically demonstrated that average-logits for the softmax-based model ensemble was better than average-token-vote in most situations.

\begin{table}[t]
    \centering
    \setlength{\tabcolsep}{1.5mm}{
    \begin{tabular}{l|cccc}
    \Xhline{1.5pt}
strategy\textbackslash{}lang & hi    & en    & zh    & fa    \\\hline
avg                      & 0.809 & 0.649 & 0.705   & 0.658 \\
avg-token-vote           & 0.833 & 0.681 & 0.732 & 0.679 \\
avg-logits               & \textbf{0.838} & \textbf{0.687}  & \textbf{0.740} & \textbf{0.685} \\\Xhline{1.5pt}
\end{tabular}}
    \caption{Results of the 5-fold cross-validation trial. ``avg'' denoted the average results of 5 models' scores. ``avg-token-vote'' represented the averagely token-vote process. ``avg-logits'' was average logits of 5 models fed into the backend softmax layer for classification.}
    \label{taby}
\end{table}
\vspace{-2mm}


\section{Conclusion}

This paper presents the implementation of the USTC-NELSLIP system submitted to the SemEval-2023 Task 2 MultiCoNER II. Different from MultiCoNER I, it has a fine-grained taxonomy which greatly increased the difficulty of the task.
The {\MODELNAME{}} method is proposed to statistically construct gazetteer and dually adapt the gazetteer network to the language model, achieving great improvements on the fine-grained NER task.
Some construction methods for gazetteers and augment data are also provided.
In future works, we will improve the gazetteer quality and apply this method to more tasks.

\section*{Acknowledgements}
  We thank anonymous reviewers for their valuable comments.

\section*{Limitations}
Although the proposed method has shown great performance for MultiCoNER, this method still can be further improved.
For instance, some textual enhancement could be adopted from Wikipedia so that the model could get stronger semantic information.
Besides, some entity types in the gazetteer has a low coverage because statistical construction is not precise enough.
In addition, since the gazetteer has a large number of irrelevant entities, denoising the gazetteer is worth studying.


\bibliography{main}

\begin{thebibliography}{27}
\expandafter\ifx\csname natexlab\endcsname\relax\def\natexlab#1{#1}\fi

\bibitem[{Ashwini and Choi(2014)}]{DBLP:journals/corr/AshwiniC14}
Sandeep Ashwini and Jinho~D. Choi. 2014.
\newblock \href {http://arxiv.org/abs/1408.0782} {Targetable named entity
  recognition in social media}.
\newblock \emph{CoRR}, abs/1408.0782.

\bibitem[{Chen et~al.(2022)Chen, Ma, Qi, Guo, Ling, and
  Liu}]{DBLP:conf/semeval/ChenMQGLL22}
Beiduo Chen, Jun{-}Yu Ma, Jiajun Qi, Wu~Guo, Zhen{-}Hua Ling, and Quan Liu.
  2022.
\newblock \href {https://doi.org/10.18653/v1/2022.semeval-1.223}
  {{USTC-NELSLIP} at semeval-2022 task 11: Gazetteer-adapted integration
  network for multilingual complex named entity recognition}.
\newblock In \emph{Proceedings of the 16th International Workshop on Semantic
  Evaluation, SemEval@NAACL 2022, Seattle, Washington, United States, July
  14-15, 2022}, pages 1613--1622. Association for Computational Linguistics.

\bibitem[{Conneau et~al.(2020)Conneau, Khandelwal, Goyal, Chaudhary, Wenzek,
  Guzm{\'{a}}n, Grave, Ott, Zettlemoyer, and
  Stoyanov}]{DBLP:conf/acl/ConneauKGCWGGOZ20}
Alexis Conneau, Kartikay Khandelwal, Naman Goyal, Vishrav Chaudhary, Guillaume
  Wenzek, Francisco Guzm{\'{a}}n, Edouard Grave, Myle Ott, Luke Zettlemoyer,
  and Veselin Stoyanov. 2020.
\newblock \href {https://doi.org/10.18653/v1/2020.acl-main.747} {Unsupervised
  cross-lingual representation learning at scale}.
\newblock In \emph{Proceedings of the 58th Annual Meeting of the Association
  for Computational Linguistics, {ACL} 2020, Online, July 5-10, 2020}, pages
  8440--8451. Association for Computational Linguistics.

\bibitem[{Craswell et~al.(2020)Craswell, Campos, Mitra, Yilmaz, and
  Billerbeck}]{DBLP:conf/cikm/CraswellCMYB20}
Nick Craswell, Daniel Campos, Bhaskar Mitra, Emine Yilmaz, and Bodo Billerbeck.
  2020.
\newblock \href {https://doi.org/10.1145/3340531.3412779} {{ORCAS:} 20 million
  clicked query-document pairs for analyzing search}.
\newblock In \emph{{CIKM} '20: The 29th {ACM} International Conference on
  Information and Knowledge Management, Virtual Event, Ireland, October 19-23,
  2020}, pages 2983--2989. {ACM}.

\bibitem[{Delobelle et~al.(2020)Delobelle, Winters, and
  Berendt}]{DBLP:conf/emnlp/DelobelleWB20}
Pieter Delobelle, Thomas Winters, and Bettina Berendt. 2020.
\newblock \href {https://doi.org/10.18653/v1/2020.findings-emnlp.292} {Robbert:
  a dutch roberta-based language model}.
\newblock In \emph{Findings of the Association for Computational Linguistics:
  {EMNLP} 2020, Online Event, 16-20 November 2020}, volume {EMNLP} 2020 of
  \emph{Findings of {ACL}}, pages 3255--3265. Association for Computational
  Linguistics.

\bibitem[{Devlin et~al.(2019)Devlin, Chang, Lee, and
  Toutanova}]{DBLP:conf/naacl/DevlinCLT19}
Jacob Devlin, Ming{-}Wei Chang, Kenton Lee, and Kristina Toutanova. 2019.
\newblock \href {https://doi.org/10.18653/v1/n19-1423} {{BERT:} pre-training of
  deep bidirectional transformers for language understanding}.
\newblock In \emph{Proceedings of the 2019 Conference of the North American
  Chapter of the Association for Computational Linguistics: Human Language
  Technologies, {NAACL-HLT} 2019, Minneapolis, MN, USA, June 2-7, 2019, Volume
  1 (Long and Short Papers)}, pages 4171--4186. Association for Computational
  Linguistics.

\bibitem[{Fetahu et~al.(2023{\natexlab{a}})Fetahu, Chen, Kar, Rokhlenko, and
  Malmasi}]{multiconer2-data}
Besnik Fetahu, Zhiyu Chen, Sudipta Kar, Oleg Rokhlenko, and Shervin Malmasi.
  2023{\natexlab{a}}.
\newblock {MultiCoNER v2: a Large Multilingual dataset for Fine-grained and
  Noisy Named Entity Recognition}.

\bibitem[{Fetahu et~al.(2021)Fetahu, Fang, Rokhlenko, and
  Malmasi}]{DBLP:conf/sigir/FetahuFRM21}
Besnik Fetahu, Anjie Fang, Oleg Rokhlenko, and Shervin Malmasi. 2021.
\newblock \href {https://doi.org/10.1145/3404835.3463102} {Gazetteer enhanced
  named entity recognition for code-mixed web queries}.
\newblock In \emph{{SIGIR} '21: The 44th International {ACM} {SIGIR} Conference
  on Research and Development in Information Retrieval, Virtual Event, Canada,
  July 11-15, 2021}, pages 1677--1681. {ACM}.

\bibitem[{Fetahu et~al.(2023{\natexlab{b}})Fetahu, Kar, Chen, Rokhlenko, and
  Malmasi}]{multiconer2-report}
Besnik Fetahu, Sudipta Kar, Zhiyu Chen, Oleg Rokhlenko, and Shervin Malmasi.
  2023{\natexlab{b}}.
\newblock {SemEval-2023 Task 2: Fine-grained Multilingual Named Entity
  Recognition (MultiCoNER 2)}.
\newblock In \emph{Proceedings of the 17th International Workshop on Semantic
  Evaluation (SemEval-2023)}. Association for Computational Linguistics.

\bibitem[{Huang et~al.(2015)Huang, Xu, and Yu}]{DBLP:journals/corr/HuangXY15}
Zhiheng Huang, Wei Xu, and Kai Yu. 2015.
\newblock \href {http://arxiv.org/abs/1508.01991} {Bidirectional {LSTM-CRF}
  models for sequence tagging}.
\newblock \emph{CoRR}, abs/1508.01991.

\bibitem[{Jia et~al.(2020)Jia, Shi, Yang, and Zhang}]{DBLP:conf/emnlp/JiaSYZ20}
Chen Jia, Yuefeng Shi, Qinrong Yang, and Yue Zhang. 2020.
\newblock \href {https://doi.org/10.18653/v1/2020.emnlp-main.518} {Entity
  enhanced {BERT} pre-training for chinese {NER}}.
\newblock In \emph{Proceedings of the 2020 Conference on Empirical Methods in
  Natural Language Processing, {EMNLP} 2020, Online, November 16-20, 2020},
  pages 6384--6396. Association for Computational Linguistics.

\bibitem[{Jiang et~al.(2020)Jiang, He, Chen, Liu, Gao, and
  Zhao}]{DBLP:conf/acl/JiangHCLGZ20}
Haoming Jiang, Pengcheng He, Weizhu Chen, Xiaodong Liu, Jianfeng Gao, and Tuo
  Zhao. 2020.
\newblock \href {https://doi.org/10.18653/v1/2020.acl-main.197} {{SMART:}
  robust and efficient fine-tuning for pre-trained natural language models
  through principled regularized optimization}.
\newblock In \emph{Proceedings of the 58th Annual Meeting of the Association
  for Computational Linguistics, {ACL} 2020, Online, July 5-10, 2020}, pages
  2177--2190. Association for Computational Linguistics.

\bibitem[{Lafferty et~al.(2001)Lafferty, McCallum, and
  Pereira}]{DBLP:conf/icml/LaffertyMP01}
John~D. Lafferty, Andrew McCallum, and Fernando C.~N. Pereira. 2001.
\newblock Conditional random fields: Probabilistic models for segmenting and
  labeling sequence data.
\newblock In \emph{Proceedings of the Eighteenth International Conference on
  Machine Learning {(ICML} 2001), Williams College, Williamstown, MA, USA, June
  28 - July 1, 2001}, pages 282--289. Morgan Kaufmann.

\bibitem[{Liu et~al.(2019)Liu, Yao, and Lin}]{DBLP:conf/acl/LiuYL19}
Tianyu Liu, Jin{-}Ge Yao, and Chin{-}Yew Lin. 2019.
\newblock \href {https://doi.org/10.18653/v1/p19-1524} {Towards improving
  neural named entity recognition with gazetteers}.
\newblock In \emph{Proceedings of the 57th Conference of the Association for
  Computational Linguistics, {ACL} 2019, Florence, Italy, July 28- August 2,
  2019, Volume 1: Long Papers}, pages 5301--5307. Association for Computational
  Linguistics.

\bibitem[{Liu et~al.(2020)Liu, Cheng, He, Chen, Wang, Poon, and
  Gao}]{DBLP:journals/corr/abs-2004-08994}
Xiaodong Liu, Hao Cheng, Pengcheng He, Weizhu Chen, Yu~Wang, Hoifung Poon, and
  Jianfeng Gao. 2020.
\newblock \href {http://arxiv.org/abs/2004.08994} {Adversarial training for
  large neural language models}.
\newblock \emph{CoRR}, abs/2004.08994.

\bibitem[{Loshchilov and Hutter(2019)}]{DBLP:conf/iclr/LoshchilovH19}
Ilya Loshchilov and Frank Hutter. 2019.
\newblock \href {https://openreview.net/forum?id=Bkg6RiCqY7} {Decoupled weight
  decay regularization}.
\newblock In \emph{7th International Conference on Learning Representations,
  {ICLR} 2019, New Orleans, LA, USA, May 6-9, 2019}. OpenReview.net.

\bibitem[{Ma et~al.(2022)Ma, Chen, Gu, Ling, Guo, Liu, Chen, and
  Liu}]{DBLP:conf/emnlp/MaCGLGL0L22}
Jun{-}Yu Ma, Beiduo Chen, Jia{-}Chen Gu, Zhenhua Ling, Wu~Guo, Quan Liu,
  Zhigang Chen, and Cong Liu. 2022.
\newblock \href {https://aclanthology.org/2022.emnlp-main.345} {Wider {\&}
  closer: Mixture of short-channel distillers for zero-shot cross-lingual named
  entity recognition}.
\newblock In \emph{Proceedings of the 2022 Conference on Empirical Methods in
  Natural Language Processing, {EMNLP} 2022, Abu Dhabi, United Arab Emirates,
  December 7-11, 2022}, pages 5171--5183. Association for Computational
  Linguistics.

\bibitem[{Malmasi et~al.(2022{\natexlab{a}})Malmasi, Fang, Fetahu, Kar, and
  Rokhlenko}]{multiconer-data}
Shervin Malmasi, Anjie Fang, Besnik Fetahu, Sudipta Kar, and Oleg Rokhlenko.
  2022{\natexlab{a}}.
\newblock {MultiCoNER: a Large-scale Multilingual dataset for Complex Named
  Entity Recognition}.

\bibitem[{Malmasi et~al.(2022{\natexlab{b}})Malmasi, Fang, Fetahu, Kar, and
  Rokhlenko}]{multiconer-report}
Shervin Malmasi, Anjie Fang, Besnik Fetahu, Sudipta Kar, and Oleg Rokhlenko.
  2022{\natexlab{b}}.
\newblock {SemEval-2022 Task 11: Multilingual Complex Named Entity Recognition
  (MultiCoNER)}.
\newblock In \emph{Proceedings of the 16th International Workshop on Semantic
  Evaluation (SemEval-2022)}. Association for Computational Linguistics.

\bibitem[{Meng et~al.(2021)Meng, Fang, Rokhlenko, and
  Malmasi}]{DBLP:conf/naacl/MengFRM21}
Tao Meng, Anjie Fang, Oleg Rokhlenko, and Shervin Malmasi. 2021.
\newblock \href {https://doi.org/10.18653/v1/2021.naacl-main.118} {{GEMNET:}
  effective gated gazetteer representations for recognizing complex entities in
  low-context input}.
\newblock In \emph{Proceedings of the 2021 Conference of the North American
  Chapter of the Association for Computational Linguistics: Human Language
  Technologies, {NAACL-HLT} 2021, Online, June 6-11, 2021}, pages 1499--1512.
  Association for Computational Linguistics.

\bibitem[{Nguyen et~al.(2016)Nguyen, Rosenberg, Song, Gao, Tiwary, Majumder,
  and Deng}]{DBLP:conf/nips/NguyenRSGTMD16}
Tri Nguyen, Mir Rosenberg, Xia Song, Jianfeng Gao, Saurabh Tiwary, Rangan
  Majumder, and Li~Deng. 2016.
\newblock \href {http://ceur-ws.org/Vol-1773/CoCoNIPS\_2016\_paper9.pdf} {{MS}
  {MARCO:} {A} human generated machine reading comprehension dataset}.
\newblock In \emph{Proceedings of the Workshop on Cognitive Computation:
  Integrating neural and symbolic approaches 2016 co-located with the 30th
  Annual Conference on Neural Information Processing Systems {(NIPS} 2016),
  Barcelona, Spain, December 9, 2016}, volume 1773 of \emph{{CEUR} Workshop
  Proceedings}. CEUR-WS.org.

\bibitem[{Rijhwani et~al.(2020)Rijhwani, Zhou, Neubig, and
  Carbonell}]{DBLP:conf/acl/RijhwaniZNC20}
Shruti Rijhwani, Shuyan Zhou, Graham Neubig, and Jaime~G. Carbonell. 2020.
\newblock \href {https://doi.org/10.18653/v1/2020.acl-main.722} {Soft
  gazetteers for low-resource named entity recognition}.
\newblock In \emph{Proceedings of the 58th Annual Meeting of the Association
  for Computational Linguistics, {ACL} 2020, Online, July 5-10, 2020}, pages
  8118--8123. Association for Computational Linguistics.

\bibitem[{Sang and Meulder(2003)}]{DBLP:conf/conll/SangM03}
Erik F. Tjong~Kim Sang and Fien~De Meulder. 2003.
\newblock \href {https://aclanthology.org/W03-0419/} {Introduction to the
  conll-2003 shared task: Language-independent named entity recognition}.
\newblock In \emph{Proceedings of the Seventh Conference on Natural Language
  Learning, CoNLL 2003, Held in cooperation with {HLT-NAACL} 2003, Edmonton,
  Canada, May 31 - June 1, 2003}, pages 142--147. {ACL}.

\bibitem[{Wang et~al.(2021)Wang, Jiang, Bach, Wang, Huang, Huang, and
  Tu}]{DBLP:conf/acl/WangJBWHHT20}
Xinyu Wang, Yong Jiang, Nguyen Bach, Tao Wang, Zhongqiang Huang, Fei Huang, and
  Kewei Tu. 2021.
\newblock \href {https://doi.org/10.18653/v1/2021.acl-long.142} {Improving
  named entity recognition by external context retrieving and cooperative
  learning}.
\newblock In \emph{Proceedings of the 59th Annual Meeting of the Association
  for Computational Linguistics and the 11th International Joint Conference on
  Natural Language Processing, {ACL/IJCNLP} 2021, (Volume 1: Long Papers),
  Virtual Event, August 1-6, 2021}, pages 1800--1812. Association for
  Computational Linguistics.

\bibitem[{Wolf et~al.(2019)Wolf, Debut, Sanh, Chaumond, Delangue, Moi, Cistac,
  Rault, Louf, Funtowicz, and Brew}]{DBLP:journals/corr/abs-1910-03771}
Thomas Wolf, Lysandre Debut, Victor Sanh, Julien Chaumond, Clement Delangue,
  Anthony Moi, Pierric Cistac, Tim Rault, R{\'{e}}mi Louf, Morgan Funtowicz,
  and Jamie Brew. 2019.
\newblock \href {http://arxiv.org/abs/1910.03771} {Huggingface's transformers:
  State-of-the-art natural language processing}.
\newblock \emph{CoRR}, abs/1910.03771.

\bibitem[{Ye and Ling(2018)}]{DBLP:conf/acl/YeL18}
Zhi{-}Xiu Ye and Zhen{-}Hua Ling. 2018.
\newblock \href {https://doi.org/10.18653/v1/P18-2038} {Hybrid semi-markov
  {CRF} for neural sequence labeling}.
\newblock In \emph{Proceedings of the 56th Annual Meeting of the Association
  for Computational Linguistics, {ACL} 2018, Melbourne, Australia, July 15-20,
  2018, Volume 2: Short Papers}, pages 235--240. Association for Computational
  Linguistics.

\bibitem[{Yu et~al.(2020)Yu, Bohnet, and Poesio}]{DBLP:conf/acl/YuBP20}
Juntao Yu, Bernd Bohnet, and Massimo Poesio. 2020.
\newblock \href {https://doi.org/10.18653/v1/2020.acl-main.577} {Named entity
  recognition as dependency parsing}.
\newblock In \emph{Proceedings of the 58th Annual Meeting of the Association
  for Computational Linguistics, {ACL} 2020, Online, July 5-10, 2020}, pages
  6470--6476. Association for Computational Linguistics.

\end{thebibliography}
\bibliographystyle{acl_natbib}



\end{document}